\documentclass[sigconf]{acmart}        
\usepackage[misc]{ifsym}
\usepackage{microtype}
\usepackage{mytables}
\usepackage{tcolorbox}
\usepackage{amsmath}
\usepackage{textcomp}    

\newcommand{\nop}[1]{}

\newcommand{\quotes}[1]{``#1''}

\let\sample\emph

\usepackage{subcaption}     

\usepackage[normalem]{ulem}
\useunder{\uline}{\ul}{}
\usepackage{multirow} 

\usepackage{listings}
\copyrightyear{2024}
\acmYear{2024}
\setcopyright{acmlicensed}\acmConference[CIKM '24]{Proceedings of the 33rd ACM International Conference on Information and Knowledge Management}{October 21--25, 2024}{Boise, ID, USA}
\acmBooktitle{Proceedings of the 33rd ACM International Conference on Information and Knowledge Management (CIKM '24), October 21--25, 2024, Boise, ID, USA}
\acmDOI{10.1145/3627673.3679791}
\acmISBN{979-8-4007-0436-9/24/10}

\begin{document}

\title{AgentRE: An Agent-Based Framework for Navigating Complex Information Landscapes in Relation Extraction}
\author{Yuchen Shi}
\email{ycshi21@m.fudan.edu.cn}
\orcid{0009-0002-6092-4304}
\affiliation{
  \institution{School of Data Science, Fudan University}
  \city{Shanghai}
  \country{China}
}

\author{Guochao Jiang}
\email{gcjiang22@m.fudan.edu.cn}
\orcid{0009-0002-3415-4473}
\affiliation{
  \institution{School of Data Science, Fudan University}
  \city{Shanghai}
  \country{China}
}

\author{Tian Qiu}
\email{tqiu22@m.fudan.edu.cn}
\affiliation{
  \institution{School of Data Science, Fudan University}
  \city{Shanghai}
  \country{China}
}

\author{Deqing Yang\textsuperscript{ \Letter}}
\email{yangdeqing@fudan.edu.cn}
\orcid{0000-0002-1390-3861}
\affiliation{
  \institution{School of Data Science, Fudan University}
  \city{Shanghai}
  \country{China}
}


\begin{abstract}
The relation extraction (RE) in complex scenarios faces some challenges such as diverse relation types and ambiguous relations between entities within a single sentence, leading to the poor performance of pure \quotes{text-in, text-out} language models (LMs). To address these challenges, in this paper we propose an agent-based RE framework, namely \emph{AgentRE}, which employs a large language model (LLM) as the agent interacting with some modules 
to achieve complex RE tasks. 
Specifically, three major modules are built in AgentRE serving as the tools to help the agent acquire and process various useful information, thereby obtaining improved RE performance. 
Our extensive experimental results upon two datasets in English and Chinese, respectively, demonstrate our AgentRE's superior performance, 
especially in low-resource scenarios. Additionally, the trajectories generated by 
AgentRE can be refined to construct a high-quality training dataset incorporating different reasoning methods, which can be used to fine-tune smaller models.\footnote{Code is available at \url{https://github.com/Lightblues/AgentRE}.} 

\end{abstract}




\maketitle


\section{Introduction} \label{sec:intro}

Relation extraction (RE) aims to transform unstructured text into structured information (relational triple), and plays a pivotal role in many downstream tasks, including semantic understanding and knowledge graph (KG) construction \cite{SciERC,MultiHead}. However, some challenges such as the diversity of relation types and the ambiguity of relations between entities in a sentence \cite{TPLinker,ETL-Span}, often hinder the models of ``\textbf{text-in, text-out}'' scheme\cite{CooperKGC,CodeKGC} from achieving effective RE. 

In recent years, large language models (LLMs) have demonstrated powerful capabilities including natural language understanding and generation, and thus been widely employed in many tasks \cite{RE-survey,LLM-RS-survey,LLM-survey-RUC}. There have been some efforts employing LLMs to achieve information extraction tasks, through converting structured extraction tasks into sequence-to-sequence tasks of natural language generation. These approaches usually adopt natural language or code to describe relation schemata \cite{InstructUIE,Code4UIE}. Despite of their advancements, these approaches are often restricted to supervised fine-tuning \cite{InstructUIE,UIE} or few-shot QA-based extraction \cite{QA4RE,Multi-Turn-QA-RE,ChatIE}, less exploring LLMs' potential in complex RE scenarios.

It is worth noting that, employing LLMs to achieve the RE tasks in complex scenarios has to face several challenges as follows:

1. \emph{How to utilize LLMs's capabilities to better leverage various significant information related to RE?} 
There exists various information, such as labelled samples, the articles and the knowledge from KGs related to the objective relations, that can be leveraged by RE models to improve RE performance. 
However, the limited context window of LLMs hinders the full utilization of comprehensive significant information. 

2. \emph{How to leverage LLMs to achieve RE effectively in specific or low-resource domains?} 
Many specific domains only have sparse data, making traditional supervised models difficult to obtain satisfactory performance. 

3. \emph{How to achieve effective RE with affordable costs?} 
Although LLMs have better performance, relatively smaller models are still considerable in practise for their affordable computational resource consumption. Thus, using the knowledge distilled from larger models to fine-tune smaller models is a reasonable way. 

Previous works \cite{agent-survey,agent-CognitiveArchitectures} have demonstrated that, the agent-based framework can endow LLMs with more capabilities such as memory, reflection and interaction with outside environment, thereby  facilitating the achievement of complex RE.
Inspired by them, 
in this paper we propose a novel agent-based framework for RE, namely \textbf{AgentRE}, which addresses the aforementioned challenges as follows. 

\figIntro

Firstly, to better leverage various significant information in complex contexts, AgentRE employs the LLM as an agent and processes the data from various sources. It utilizes some tools such as retrieval and memory module to aid the agent's reasoning process. For instance, as illustrated in Figure \ref{fig:intro}, unlike conventional \quotes{text-in, text-out} LMs relying on single-round input-output to achieve RE, AgentRE engages in multiple rounds of interaction and reasoning. This approach enables the utilization of a broader spectrum of information sources for extraction tasks, avoiding the limitations in single-round extraction. 

Secondly, facing the situations of low-resource, AgentRE can make dynamic summarizations and reflections throughout the extraction process with the help of the LLM's reasoning and memory capability. As a result, AgentRE is adept at continual learning, improving its extraction capability through an ongoing process of summarizing experiences and accumulating knowledge.

Finally, we introduce a method for converting the reasoning trajectories of AgentRE into high-quality data, which encompass various reasoning strategies such as direct generation, step-by-step extraction, and CoT (Chain-of-Thought) based extraction. The enriched data can be utilized to fine-tune relatively small models, guiding them to dynamically select different extraction methods (as discussed in \cite{FireAct}), thereby enhancing the small models' extraction performance.

In summary, the main contributions of this paper include:
    
    1. We propose an agent-based RE framework \emph{AgentRE}, in which the agent can explore and collect more significant information to improve RE, with the retrieval, memory and extraction modules.
    
    2. Our extensive experiments on two datasets in English and Chinese not only validate AgentRE's state-of-the-art (SOTA) performance in low-resource RE tasks, but also verify the effectiveness of each module built in AgentRE.

    3. By utilizing the reasoning trajectories of the agent in AgentRE, the refined records can be utilized to construct a dataset incorporating diverse reasoning methods. Through distillation learning, the reasoning-based extraction capabilities can be transferred from large models to relatively small models, to achieve satisfactory RE with affordable costs.

\section{Related Work} \label{sec:related}

\subsection{LLM-based Information Extraction} \label{subsec:llm-ie}
Recent studies \cite{InstructUIE, Code4UIE, ChatIE, CodeKGC} have explored using LLMs for information extraction (IE). 
The research can be categorized into two groups. The first group focuses on LLMs designed for specific IE tasks, such as named entity recognition (NER) \cite{UniNER}, relation extraction (RE) \cite{QA4RE}, and event extraction (EE) \cite{ClarET}. These models often perform better but require separate fine-tuning for each task. 
The second group aims to handle multiple IE tasks with a single model, creating a universal extraction model \cite{InstructUIE, Code4UIE, UIE}. This approach uses a unified method with designed prompts to address various tasks, enhancing generalization but sometimes underperforming on specific tasks \cite{LLMIE-survey}.

Furthermore, CooperKGC \cite{CooperKGC} has tried to utilize agents to tackle diverse IE subtasks. It emphasizes information interaction among multiple agents, using individual agents for different subtask. In contrast, our paper explores various types of information sources that could be utilized in IE tasks, with a stronger focus on leveraging agent memory and reasoning to accomplish extraction in complex scenarios.

\subsection{LLM-based Agent} \label{subsec:llm-agent}

In recent years, LLM-based agents have gained significant attention. LLMs demonstrate strong task-solving and reasoning capabilities for both real and virtual environments. These abilities resemble human cognitive functions, enabling these agents to perform complex tasks and interact effectively in dynamic settings.

\noindent\emph{Plannning}: It involves the ability to strategize and prepare for future actions or goals. AUTOACT \cite{AUTOACT} introduces an automatic agent learning framework for planning that does not rely on large-scale annotated data and synthetic trajectories from closed-source models (e.g., GPT-4).

\noindent\emph{Tool Use}: This is the capacity to employ objects or instruments in the environment to perform tasks, manipulate surroundings, or solve problems. KnowAgent \cite{KnowAgent} introduces a novel approach designed to enhance the planning capabilities of LLMs by incorporating explicit action knowledge.

\noindent\emph{Embodied Control}: It refers to an agent's ability to manage and coordinate its physical form within an environment. This encompasses locomotion, dexterity, and the manipulation of objects. RoboCat \cite{RoboCat} introduces a visual goal-conditioned decision transformer capable of consuming action-labeled visual experience.

\noindent\emph{Communication}: It is the skill to convey information and understand messages from other agents or humans. Agents with advanced communication abilities can participate in dialogue, collaborate with others, and adjust their behaviour based on the communication received. \citet{BootstrappingLT} introduce an automated way to measure the partial success of a dialogue, which collects data through LLMs engaging in a conversation in various roles.

In this paper, our proposed AgentRE is built based an agent interacting with the environment, which primarily utilizes the capabilities of LLMs to achieve the RE in complex scenarios. 
\section{Proposed Method} \label{sec:model}

\figFramework       

\subsection{Overview} \label{subsec:framework}
The overview of our proposed framework is illustrated in Figure \ref{fig:framework}(a), where the core LLM-based agent plays the important role of reasoning and decision-making. 
The three modules around the agent, i.e., the \emph{retrieval} module, \emph{memory} module, and \emph{extraction} module, serve as the tools to aid the agent on acquiring and processing information. 
We briefly introduce the functions of the three modules as follow.

     \paragraph{Retrieval Module} It maintains relatively static knowledge to facilitate storing and retrieving information, including the annotated samples from the training dataset and the related information such as annotation guidelines. 
    
    \paragraph{Memory Module} It maintains relatively dynamic knowledge, including \emph{shallow memory} for recording extraction results and \emph{deep memory} for summarizing and reflecting on historical actions. Our framework records and utilizes the extraction experiences by reading from and writing to the memory module.
    
     \paragraph{Extraction Module} It extracts structured information (triples) from the input text with various reasoning methods, based on the information provided by the retrieval and memory module.

Next, we introduce the design details of all modules in AgentRE.

\subsection{Retrieval Module} \label{subsec:retrieval}
The retrieval module in our framework serves as a critical component to source relevant samples from existing datasets and supplementary knowledge from various resources, and thus helps the extraction module achieve RE task. The retrievable data may be extensive and diverse, which, for the purpose of clarity, is categorized into two main types in this paper. 
    
    1. Labelled data with a clear input-output relationship $x \rightarrow y$, which can be organized into the context of the LLM as the few-shot examples, helping the model quickly understand the input-output relationship of the current task.
    
    2. Other relevant information, such as relation descriptions, annotation guidelines, and even external knowledge in encyclopedia. By injecting them as aside information into the context of the LLM, they can assist the model on understanding the extraction task.\footnote{For a fair comparison with existing models, in our experiments our AgentRE does not leverage external web knowledge such as encyclopedia sites. However, existing work \cite{AutoKG} has conducted the experiments in such a setting.}

\begin{figure}
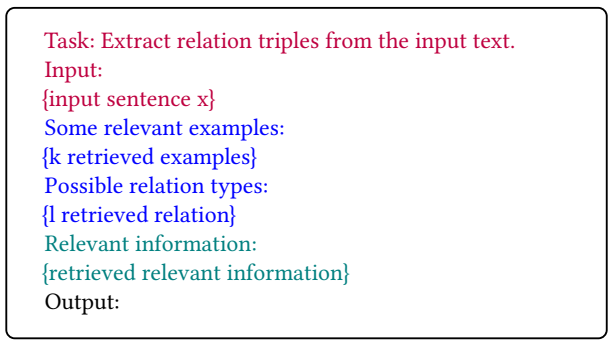
   
\centering
\begin{tcolorbox}[colback=white, colframe=black, boxrule=0.25mm, width=0.45\textwidth]
    \textcolor{purple}{Task: Extract relation triples from the input text.}\\
    \textcolor{purple}{Input:}\\
    \textcolor{purple}{\{input sentence x\}}\\
    \textcolor{blue}{Some relevant examples: }\\
    \textcolor{blue}{\{k retrieved examples\}}\\
    \textcolor{blue}{Possible relation types: }\\
    \textcolor{blue}{ \{l retrieved relation\}}\\
    \textcolor{teal}{Relevant information: }\\
    \textcolor{teal}{ \{retrieved relevant information\}}\\
    \textcolor{black}{Output:}
\end{tcolorbox}

\caption[Context format of prompts]{The prompt template for the retrieval module.} \label{fig:agentRE-io}
\Description{This figure illustrates the structured format of prompts used by the retrieval module for the task of extracting relation triples from text. The prompt is divided into sections highlighted in different colors: purple for the task description and input sentence, blue for examples and possible relation types, teal for relevant information, and black for the output section. Each section is clearly labeled to guide the model in processing the input text and generating the appropriate output.}
\end{figure}

To effectively manage and utilize these two types of data, we introduce two specific retrieval modules: the \emph{sample retrieval module} and the \emph{relevant information retrieval module}. 
Once informative labelled data and other pertinent information are acquired, the retrieval module can leverage these insights. A straightforward approach is to concatenate them into prompts, thereby assimilating this beneficial information. The template of these prompts is depicted in Figure \ref{fig:agentRE-io}. It is worth mentioning that the extraction module may adopt various reasoning methods other then straightforward prompting, as detailed in Section \ref{subsec:extraction}.

\subsubsection{Sample Retrieval} \label{subsub:agent-retrieval-example}
The sample retrieval module, as shown in the lower part of Figure \ref{fig:framework}(b), encodes the current text into an embedding with an encoder. It then calculates the similarities between the samples in the training dataset to retrieve the samples similar to the current text. For instance, for the sentence \sample{``On May 9th, Nobel laureate and writer Mo Yan delivered a speech in Beijing.''}, the sample retrieval module can retrieve relevant samples from the training dataset through embedding matching, such as the text \sample{``When the newly minted Nobel Prize in Literature, British novelist Kazuo Ishiguro, found himself...''} with its corresponding label (relational triple) as \sample{(Kazuo Ishiguro, award, Nobel Prize in Literature)}.

Specifically, the sample retrieval module includes a pretrained text encoder for converting input text into embedding, and an embedding retriever for retrieving the samples similar to the input text from the training dataset. Given the current input text $x$, it is encoded into an embedding $\mathbf{e}_x$, just like all samples $\{t_1,t_2,...,t_N\}$ in the training dataset, as follows:
\begin{align}
    \mathbf{e}_x &= \text{Encoder}(x),\\
    \mathbf{e}_{t_i} &= \text{Encoder}(t_i), \quad i=1,2,...,N.
\end{align}
For all sample embedding set $\mathbf{E} = \{\mathbf{e}_{t_1},\mathbf{e}_{t_2},...,\mathbf{e}_{t_N}\}$ constructed from the training data, the similarity between the input text embedding $\mathbf{e}_x$ and each sample embedding $\mathbf{e}_{t_i}$ can be calculated as cosine similarity, thus to obtain a similarity vector $\mathbf{s} = \{s_1,s_2,...,s_N\}$ where $s_i = \operatorname{cosine}(\mathbf{e}_x,\mathbf{e}_{t_i})$. Based on these similarity scores, the $k$ most similar samples to the input text are retrieved. In fact, such an embedding retrieval process can be implemented through a standard retriever as
\begin{equation}
    \{t_{i_1},t_{i_2},...,t_{i_k}\} = \operatorname{EmbeddingRetriever}(\mathbf{e}_x,\mathbf{E},k),
\end{equation}
where $\mathbf{E}$ is the embedding set and $\{i_1,i_2,...,i_k\}$ represents the retrieved samples' positions in training dataset.

Additionally, when facing a large number of relation types, the extraction process might be decomposed into two distinct phases: identifying potential relation types presenting in the sentence, and then conducting the extraction based on these identified candidate relation types. The process of retrieving candidate relation types is represented by the dashed arrow in Figure \ref{fig:framework} (b).
A feasible approach for this retrieval is to develop a classifier trained on the dataset to predict the relations most likely to be found in the given text. Furthermore, the task of retrieving relation types can also be achieved using the inferential capabilities of LLMs, as discussed in Section \ref{subsec:extraction}.

\subsubsection{Relevant Information Retrieval} \label{subsub:agent-retrieval-related}
The relevant information retrieval module, as shown in the upper part of Figure \ref{fig:framework}(b), is used to retrieve knowledge related to the given sentence. Compared to the embedding retrieval method used in Sample Retrieval, this module employs a variety of retrieval methods mixing vectors and entities to combine precise matching and fuzzy semantic matching.

For example, for the same sentence \sample{``On May 9th, Nobel laureate and writer Mo Yan delivered a speech in Beijing.''}, besides leveraging the sentence's representation, this module also identifies potential entities in the sentence, such as \sample{Mo Yan}, \sample{Nobel Prize} and \sample{Beijing}, and retrieves related knowledge using these entities. 
Additionally, based on the entity \sample{Nobel Prize}, explanatory information about the candidate relation type \sample{award}, including the definition of the head and tail entities of this relation type and detailed explanations, can be retrieved together from the annotation guidelines.

Formally, the relevant information retrieval module includes the preprocessing part of extracting key information or constructing embeddings, and several retrievers for retrieving information related to the input text.
In the preprocessing part, besides the text encoder, there is also an Entity Recognizer for identifying all potential entities in the input text as
\begin{equation}
    \{c_1,...,c_{C_x}\} = \operatorname{EntityRecognizer}(x),
\end{equation}
where $C_x$ is the number of entities identified in the input text $x$.
In the retriever part, various methods can be used to retrieve related knowledge from different data sources, such as retrieving the attributes and relations of the entities from knowledge graph, retrieving explanatory information about the relations from annotation guidelines, or even retrieving related knowledge from external encyclopedias.

Besides the embedding-based retriever introduced above, here we introduce an entity-based retriever for retrieving knowledge related to the input text from existing KG. It mainly includes Entity Linking and Entity Property Retrieval parts. Given a candidate entity mention $c_i$, we have
\begin{align}
    e_i &= \text{EntityLinking}(c_i),\\
    \{t_i^1,t_i^2,...,t_i^{T_i}\} &= \text{EntityPropertyRetrieval}(e_i),
\end{align}
where $e_i$ is the entity linked by the entity linker from mention $c_i$, and $\{t_i^1,t_i^2,...,t_i^{T_i}\}$ represents the triples related to entity $e_i$ in the KG.

\subsection{Memory Module} \label{subsec:memory}
The roles of the memory module in AgentRE include dynamically utilizing existing knowledge during the extraction process, reflection and summarization, which helps AgentRE better achieve subsequent extraction tasks. Mimicking the human brain, the model's memory can be divided into \emph{shallow memory} and \emph{deep memory}. 

\subsubsection{Shallow Memory} \label{subsub:agent-memory-shallow}
Shallow memory refers to the preliminary records of extraction experiences. For example, as illustrated in Figure \ref{fig:framework}(c), for the sentence \sample{``The Musesum is located in Northeast Gaomi Township, Mo Yan's hometown.''}, the model's extraction results are \sample{(Mo Yan, place\_of\_birth, Northeast Gaomi Township)} and \sample{(Musesum, located\_at, Northeast Gaomi Township)}. The first triple is correct but the second triple is marked as incorrect, due to the unclear referent of the mention \sample{Musesum}. 
In shallow memory, by recording the correct and incorrect results, the model can use them as the references in subsequent extractions. This process can be understood as the lessons learned from previous experiences. 
Specifically, the model adds a new record in \textbf{correct memory} and \textbf{incorrect memory}, respectively.

Formally, for an input sentence $x$, the extraction module generates $M$ triples, denoted as $\hat{Y} = \{y_1,y_2,...,y_M\} = \text{TripleExtractor}(x)$, where $y_i = (h_i,r_i,t_i)$ represents the $i$-th triple. After verifying each triple denoted as $\text{verify}(y_i)$, the correct triple set $Y_{\text{correct}} = \{y_i | y_i \in \hat{Y},\text{verify}(y_i) = \text{True}\}$ and the incorrect triple set $Y_{\text{wrong}} = \{y_i | y_i \in \hat{Y},\text{verify}(y_i) = \text{False}\}$ are obtained. Then, they are added into the memory component $\mathcal{M}_{Correct}$ or $\mathcal{M}_{Wrong}$ as 
\begin{align}
    \mathcal{M}_{Correct} &= \mathcal{M}_{Correct} \cup Y_{\text{correct}},\\
    \mathcal{M}_{Wrong} &= \mathcal{M}_{Wrong} \cup Y_{\text{wrong}}.
\end{align}

\subsubsection{Deep Memory} \label{subsub:agent-memory-deep}
Deep memory includes the reflections and updates to historical memories, as shown in Figure \ref{fig:framework}(c).
In deep memory, AgentRE can \emph{update} long-term memories based on correct results and \emph{reflect} on incorrect ones. Taking the example shown in Figure \ref{fig:framework}(c), given current correct extraction result, AgentRE updates its memory on entity \sample{Mo Yan} from \sample{``Mo Yan, a famous writer, was born on February 17, 1955. His real name is Guan Moye''} to a new one \sample{``Mo Yan, a famous writer, was born on February 17, 1955, in Northeast Gaomi Township. His real name is Guan Moye.''}.
Moreover, for incorrect results, AgentRE performs {reflection}. For example, given an incorrect extraction result and relevant annotation guidelines, it generates the reflection text \sample{``Incomplete entities, such as \sample{Musesum}, should not be extracted according to the annotation guidelines''}. Thus, if the next input text is \sample{``The Musesum, named after the most influential contemporary writer and scholar Mr. Wang Meng...''}, AgentRE can avoid similar errors by referring to previous reflections.

Formally, given an input sentence $x$ and its correct extraction results $Y_{\text{correct}}$, AgentRE leverages each record (triple) $y_i \in Y_{\text{correct}}$  
to update the deep memory $\mathcal{M}_{Deep}$ as
\begin{equation}
    \mathcal{M}_{Deep} = \operatorname{UpdateDeepMemory}(\mathcal{M}_{Deep},y_i). \label{eq:agent-memory-deep}
\end{equation}
The update operation $\operatorname{UpdateDeepMemory}(,)$ includes the following three steps:
\begin{align}
    m_i &= \text{MemoryRetrieval}(\mathcal{M}_{Deep},y_i),\\
    m'_i &= \text{MemoryUpdate}(m_i,y_i),\\
    \mathcal{M}_{Deep} &= \mathcal{M}_{Deep} \setminus \{m_i\} \cup \{m'_i\}.
\end{align}
Here, $m_i$ and $m'_i$ respectively represent the retrieved original memory and the updated memory. It should be noted that when the retrieved memory is empty, i.e., no related description is found, the model directly summarizes and adds the correct result into the deep memory.

For incorrect extraction results $Y_{\text{wrong}}$, the model reflects on each record $y_j \in Y_{\text{wrong}}$ and records the reflection outcome in the reflection memory as below,
\begin{align}
    r_j &= \operatorname{Reflection}(y_j),\\
    \mathcal{M}_{Ref} &= \text{UpdateRefMemory}(\mathcal{M}_{Ref}, r_j),
\end{align}
where $r_j$ is the reflection result for the incorrect record $y_j$, and $\mathcal{M}_{Ref}$ denotes the reflection memory. Operation $\operatorname{UpdateRefMemory}(,)$ includes recalling and updating related reflection memories, similar to the update operations for deep memory in Equation \ref{eq:agent-memory-deep}.

\subsection{Extraction Module} \label{subsec:extraction}
We now present the overall extraction pipeline of extraction module in AgentRE. It adopts an interactive process similar to ReAct \cite{ReAct}, engaging in multiple rounds of \emph{Thought, Action, Observation}, as illustrated in Figure \ref{fig:framework}(d).

In this context, the retrieval and memory module are uniformly considered as the external tools used by the agent. As a series of APIs, the agent is provided with the tool name, input parameters when using these tools, and then receives the results. It allows the agent to dynamically decide \emph{whether to call tools, which tools to call, and how to call them}.

For instance, still consider the sentence in Figure \ref{fig:framework}(d)\sample{``On May 9th, Nobel laureate and writer Mo Yan delivered a speech in Beijing.''}. In the first round, the agent identifies the potential relation types and then chooses to call the \emph{SearchAnnotation} API to obtain relevant information. In the second round, the agent uses the \emph{SearchKG} API to retrieve existing knowledge about \emph{Mo Yan}. Finally, after gathering sufficient information, the agent executes the \emph{Finish} action to return the extraction results.

It is important to note that, as shown in Figure \ref{fig:framework}(d), during extraction process, AgentRE may not always follow a complete multi-round ReAct interactions. Instead, it dynamically selects the appropriate extraction method based on the complexity of the input text. For example, it may use \emph{Direct} extraction where the predicted relational triples are output directly from the input text, or \emph{Staged} extraction where the relation types are first filtered, followed by the extraction of triples, or \emph{Chain-of-Thought} (CoT) extraction where the final extraction results are generated step-by-step.

\subsection{Distillation for Smaller Models} \label{sub:agent-distillation}
In the real-world applications, employing LLMs with robust reasoning capabilities as agents to achieve extraction tasks, has to face the problem of high costs. On the other hand, (relatively) smaller large language models (SLLMs) often exhibit comparatively weaker reasoning abilities. To bridge this gap, we introduce a distillation learning approach that leverages the historical reasoning trajectories of larger models to guide the learning of smaller models.

Prior research \cite{FireAct} has shown that applying diverse reasoning strategies to different types of problems can significantly improve a model's problem-solving versatility. For instance, in the context of RE tasks, straightforward relations that are explicitly mentioned in the text can be directly inferred to produce structured outputs. For the sentences encapsulating more complex relations, employing a CoT-based reasoning approach can guide the model through a step-by-step process towards the final result, thereby minimizing errors. 
Our AgentRE's reasoning framework, as described above, effectively employs tailored reasoning methodologies for varied scenarios through the agent. To endow SLLMs with similar capabilities while simplifying the reasoning process, we propose to distill more simplified rationales from AgentRE's historical reasoning trajectories, which are utilized to direct the learning of smaller models.

Formally, the sequence of thought, action and observation generated by AgentRE can be encapsulated into the following reasoning trajectory as
\begin{equation}
    P = \left\{p_j= (t_j,a_j,o_j)\right\}_{j=1}^{|P|},
\end{equation}
where $t_j$ is the thought in the $j$-th iteration, $a_j$ denotes the action taken, and $o_j$ represents the observation, with the sequence extending over $|P|$ iterations. 
Integrating the reasoning trajectory with the input text and the accurate extraction results, allows the LLM to summary a more succinct rationale as
\begin{equation}
    \{r_i,y_i\} = \text{Summarize}(P,x_i,y_i),
\end{equation}
where $r_i$ represents the summarized rationale, and $y_i$ represents the correct extraction result. Such rationales can serve as the learning objectives for SLLMs, guiding their learning through supervised learning.

The accumulated extraction results with the rationales can be used to generate a novel training dataset ${{D}'} = \{(x_i,r_i,y_i)\}_{i=1}^{N}$, where $N$ is the total sample number. This dataset enriches the original training dataset ${D} = \{(x_i,y_i)\}_{i=1}^{N}$ with the agent's distilled reasoning experiences, incorporating adaptive reasoning strategies. The objective of distillation learning with this enriched dataset is to empower SLLMs to select the most fitting reasoning approach based on the nuances of the input sentence. This learning process (supervised fine-tuning) can be formalized as
\begin{equation}
 \theta'_{SLLM} = \text{SFT}(\theta_{SLLM},{{D}'}),
\end{equation}
where $\theta_{SLLM}$ and $\theta'_{SLLM}$ denote the initial and fine-tuned parameter set of the SLLM, respectively. 
\section{Experiments} \label{sec:exp}

\subsection{Dataset Description} \label{sub:agent-dataset}
We have conducted extensive experiments to validate the effectiveness of AgentRE on the following two datasets.

\textbf{DuIE} \cite{DuIE}\footnote{\url{https://ai.baidu.com/broad/download}.} is the largest Chinese RE dataset, comprising 48 predefined relation types. Besides traditional simple relation types, it also includes complex relation types involving multiple entities. The annotated corpus was sourced from Baidu Baike, Baidu Information Stream, and Baidu Tieba texts, encompassing 210,000 sentences and 450,000 relations.

\textbf{SciERC} \cite{SciERC}\footnote{\url{https://nlp.cs.washington.edu/sciIE/}.} is an English dataset for NER and RE in the scientific domain. The annotated data were derived from the \emph{Semantic Scholar Corpus}, covering abstracts of 500 articles. The SciERC dataset includes 8,089 entities and 4,716 relation records in total, with an average of 9.4 relations per document.

\subsection{Comparison Models} \label{sub:agent-compare}

We compared our AgentRE with several LLM-based IE models/frameworks in our experiments as follows. 

1) \textbf{ChatIE} \cite{ChatIE} introduces a zero-shot IE approach through the dialogue with ChatGPT, framing zero-shot IE as multi-turn question-answering. It first identifies possible relation types, and then extracts relational triples based on these types. 

2) \textbf{GPT-RE} \cite{GPT-RE} employs a task-aware retrieval model in a few-shot learning framework, incorporating CoT for automatic reasoning, addressing the issues of instance relevance and explanation in input-label mapping. 

3) \textbf{CodeKGC} \cite{CodeKGC} uses Python classes to represent structural schemata of relations, enhancing extraction with reasoning rationales. 

4) \textbf{CodeIE} \cite{CodeIE} transforms IE tasks into codes, leveraging LLMs' code reasoning capabilities. 

5) \textbf{UIE} \cite{UIE} introduces a structured encoding language for text-to-structured output generation, which is used for pretraining T5 model\cite{T5}. 

6) \textbf{USM} \cite{USM} proposes a unified semantic matching framework for IE with structured and conceptual abilities, which is built based on RoBERTa \cite{RoBERTa}. 

7) \textbf{InstructUIE} \cite{InstructUIE} applies instruction-based fine-tuning on Flan-T5 \cite{Flan-T5} for enhanced task generalizability.

In brief, ChatIE and CodeKGC utilize zero-shot learning with LLMs, while CodeIE, CodeKGC and GPT-RE adopt few-shot approaches. UIE, USM and InstructUIE adopt supervised fine-tuning (SFT). Notably, GPT-RE was also fine-tuned on larger models like \emph{text-davinci-003} for specific tasks, which is cost-intensive.

\subsection{Overall Results and Implementation Details} \label{sub:agent-overall-results}

The overall experimental results are listed in Table \ref{tab:AgentRE-main}. Here we only use F1 score as the metric for the alignment with previous papers.
For the baseline models, we endeavored to directly cite the scores from their original publications or reproduced the results by using their published models and source codes. 
Moreover, to ensure the fairness of our experimental comparisons, we predominantly utilized the same backbone LLM, e.g., \emph{gpt-3.5-turbo}. For the methods employing different backbone models, we have included their original results and supplemented them with the results obtained by using \emph{gpt-3.5-turbo} as the backbone model, as the italic scores in the table.\footnote{For CodeKGC, due to its reliance on the now-deprecated \emph{text-davinci-003} model, the replication on DuIE was not feasible. However, we have added the results based on \emph{gpt-3.5-turbo}. Furthermore, for USM and GPT-RE-FT, which necessitate fine-tuning, their non-public model availability precluded the replication on DuIE. }

\tableAgentREMain

Table \ref{tab:AgentRE-main} is divided into three parts based on different experimental paradigms: zero-shot learning (\emph{ZFL}), few-shot learning (\emph{FSL}), and supervised fine-tuning (\emph{SFT}) settings. 
The second column lists the backbone models used in these methods. 
For the methods under SFT setting, 
which can be roughly divided into three categories as below based on the size of the model parameters.
1) The \emph{T5-v1.1-large}\footnote{\url{https://github.com/google-research/text-to-text-transfer-transformer}.} used by UIE and the \emph{RoBERTa-Large}\footnote{\url{https://github.com/facebookresearch/fairseq/tree/main/examples/roberta}.} used by USM have parameter sizes of 0.77B and 0.35B, respectively. 2) The \emph{Flan-T5}\footnote{\url{https://github.com/facebookresearch/llama}.} used by InstructUIE and the \emph{Llama-2-7b}\footnote{\url{https://github.com/google-research/FLAN}.} used by AgentRE-SFT have parameter sizes of approximately 11B and 7B, respectively. 
3) The \emph{gpt-3.5-turbo} used by GPT-RE-SFT has the parameter size of approximately 175B. 
According to the experimental results, we have the following conclusions.

    1. In ZSL group, \emph{ChatIE-multi} outperforms \emph{ChatIE-single}, demonstrating the effectiveness of multi-turn dialogues. \emph{AgentRE-ZSL}'s superior performance indicates its efficient use of auxiliary information.
    
    2. In FSL group, \emph{CodeKGC-FSL} surpasses dialogue-based \emph{ChatIE}, and \emph{GPT-RE} matches its performance, highlighting the benefits of structured reasoning and precise sample retrieval. \emph{AgentRE-FSL} notably outperforms the SOTA models, demonstrating its superior utilization of labelled data and auxiliary information.
    
    3. Under SFT setting, fine-tuning smaller models like \emph{UIE} and \emph{USM} yields better results than the baseline models but falls short of \emph{AgentRE-FSL}. \emph{AgentRE-SFT} significantly outperforms \emph{InstructUIE}, evidencing the effectiveness of the distillation learning in AgentRE. However, \emph{GPT-RE-SFT} achieves the best performance on SciERC, albeit with higher training costs due to its large model size and API-based training on \emph{text-davinci-003}.

\subsection{Ablation and Parameter Tuning Study} \label{sub:agent-ablation}
This section displays the results of ablation and parameter tuning study, focusing on the impacts of AgentRE's retrieval and memory module on RE performance.

\subsubsection{Overall Ablation Study} \label{subsub:agent-ablation-overall}
\tableAgentREAblationOverall
The ablation study examines the performance of AgentRE under different settings: without the retrieval module (\emph{AgentRE-w/oR}), without the memory module (\emph{AgentRE-w/oM}), and lacking both (\emph{AgentRE-w/oRM}). 
The results in Table\ref{tab:AgentRE-ablation-overall}, reveal a significant underperformance of \emph{AgentRE-w/oRM}, underscoring the essential roles of both modules. \emph{AgentRE-w/oR} and \emph{AgentRE-w/oM} exhibit better performance than \emph{AgentRE-w/oRM}, verifying the value of adding the memory and retrieval module independently. Notably, the full framework AgentRE integrating both modules, achieves the best performance, demonstrating the synergistic effect of combining retrieval capabilities for accessing similar samples and the memory for capitalizing on previous extractions. 

\subsubsection{Analysis of Retrieval Module} \label{subsub:agent-ablation-retrieval}
\tableAgentREAblationRetrievalMethod
Overall, the variables affecting the retrieval module's effects mainly include the models used for data representation and retrieval and the content available for retrieval. 

\noindent\textbf{Retrieval Model}: Our experiments evaluated several retrieval methods against the baseline approach, i.e., \emph{Random}, in which $k$ labelled samples are chosen at random. These evaluated methods include statistical techniques such as TF-IDF \cite{TF-IDF} and BM25 \cite{BM25}, as well as embedding-based approaches like SimCSE \cite{SimCSE} and BGE \cite{BGE}. These two schemes employ statistical metrics and vector similarity, respectively, to fetch labelled the samples similar to the given sentence. For implementing TF-IDF and BM25, we utilized the \emph{scikit-learn}\footnote{\url{https://scikit-learn.org/stable/}} and \emph{Rank-BM25}\footnote{\url{https://github.com/dorianbrown/rank_bm25}} packages, with Chinese word segmentation performed using \emph{Jieba}\footnote{\url{https://github.com/fxsjy/jieba}}. The embedding-based models were facilitated through the \emph{SimCSE}\footnote{\url{https://github.com/princeton-nlp/SimCSE}} package and the \emph{BGE}\footnote{\url{https://github.com/FlagOpen/FlagEmbedding}} project. In this set of experiments, the focus was solely on labelled samples, disregarding other relevant information, and the number of retrieved samples was fixed at $k=5$.

The results in Table \ref{tab:AgentRE-ablation-retrieval-method} demonstrate that both statistical and embedding-based methods significantly surpass the random retrieval baseline. This indicates the effectiveness of retrieving labelled samples more closely aligned with the input text in aiding the model's decision-making process, thereby improving its extraction accuracy. Among the evaluated models, BGE showed superior performance on both datasets and was therefore selected for the retrieval module in subsequent experiments.

\tableAgentREAblationRetrievalContent

\noindent\textbf{Content for Retrieval}: Following the backbone model selection for the retrieval module, we delved into the impact of various types of available information for retrieval. As outlined in Section \ref{subsec:retrieval}, this information falls into two main categories: labelled samples and unlabelled relevant information, the latter including annotation guidelines and entity-related KG information.

Table \ref{tab:AgentRE-ablation-retrieval-content} lists the experimental results, where \emph{None} and \emph{AgentRE-w/oM} denote the variants without and only with the full retrieval module, respectively. Additionally, \emph{-samples}, \emph{-doc}, and \emph{-KG} indicate the variants without the labelled sample retrieval, annotation guidelines retrieval, and KG retrieval components, respectively. The results justify that omitting any type of information degrades AgentRE's performance, with the removal of labelled samples (\emph{-samples}) exerting the most significant impact.

In essence, this analysis emphasizes the pivotal roles that both retrieval methodologies and the scope of retrieval content enhance AgentRE's performance. The capabilities of effectively retrieving samples and integrating a broad spectrum of pertinent information are crucial for augmenting AgentRE's extraction proficiency.

\subsubsection{Analysis of Memory Module} \label{subsub:agent-ablation-memory}
\figMemoryCurve
To evaluate the impact of the memory module on RE performance, we examined the F1, Recall, and Precision scores of AgentRE with varying memory configurations on the DuIE dataset as training data quantity increased, as depicted in Figure \ref{fig:AgentRE-ablation-memory} where the X-axis of the figure is the number of training samples. The compared models include \emph{AgentRE-w/oM} (without the memory module), \emph{AgentRE-wM} (with shallow memory as described in Section \ref{subsub:agent-memory-shallow}), and \emph{AgentRE-wM+} (integrating both shallow and deep memory). 
The models with memory modules leverage both input samples and historical extraction records, unlike their memory-less counterpart. Each model began with an identical set of 200 randomly selected labelled samples for the retrieval module.

The experimental results revealed the following insights:

1) The models incorporating memory module (\emph{AgentRE-wM} and \emph{AgentRE-wM+}) outperform the memory-less variant in all metrics, underscoring the memory module's beneficial impact on extraction accuracy.

2) Performance scores for the models with memory modules improve that as more data was introduced, indicating effective utilization of past extraction experiences for dynamic learning.

3) \emph{AgentRE-wM+} demonstrated superior performance over \emph{AgentRE-wM} with increased data input, suggesting that a comprehensive approach to memory, beyond mere individual sample tracking, can further enhance extraction capabilities.

\subsection{Low-Resource Scenario} \label{sub:agent-low-resource}
\tableAgentRELowResource

We also investigated the impact of varying labelled data quantity on extraction performance by sampling different amounts ($N = 0, 10, 100, 1000$) of samples from DuIE. In this study we compared two methods: \emph{AgentRE} integrating retrieval and memory modules, and the basic in-context learning (\emph{ICL}) model employing sample retrieval similar to GPT-RE.

Table \ref{tab:AgentRE-low-resource} lists the relevant results from which we find:

1) The \emph{ICL} model's performance is highly dependent on the quantity of available training samples, with F1 scores of $34.40\%$ and $44.91\%$ at $N=10$ and $N=100$, respectively. 
It highlights the model's limitations in low-resource scenarios, where its dependence on sample retrieval for ICL, without leveraging other pertinent information, adversely affects its extraction capabilities.

2) AgentRE consistently outperforms the ICL model across all data quantities, particularly at extremely low data availabilities ($N=0, 10$). This suggests AgentRE's superior performance on leveraging the LLM for interaction and reasoning, thus more effectively utilizing available information for enhanced RE.

3) Both models exhibit performance gains with increasing $N$, affirming that additional labelled data promotes model performance by providing more relevant training samples. 

\subsection{Fine-Tuning Study} \label{sub:exp-agent-distillation}
\tableAgentREDistillation

In this subsection, we verify the effectiveness of the distillation method based on historical reasoning rationales introduced in Section \ref{sub:agent-distillation}.
When fine-tuning SLLMs, a straightforward approach is to input the sentence $x$ directly into the model, allowing it to output the predicted triples $\hat{Y}$. 
The original training data in this manner is denoted as $D$, and the dataset $D^{\prime}$ is derived from summarizing the agent's historical reasoning trajectories. 
By comparing the performance of models trained on these two different datasets, we explore the effectiveness of distillation learning. 

Specifically, $D$ includes 10,000 samples from DulE's training set, while $D^{\prime}$ contains reasoning rationales and 1,000 samples. 
In addition to comparing the models trained separately on each dataset, we also considered sequential fine-tuning on both datasets, denoted as $D+D^{\prime}$. This approach involves the initial training on the larger dataset $D$ followed by further fine-tuning on $D^{\prime}$. In all experiments, models are trained for 3 epochs on each dataset.

Parameter-efficient fine-tuning was performed using the LoRA\cite{LoRA} method, with the low-rank matrix dimension set to $r=8$, the scaling factor set to $\text{alpha}=16$, and the dropout rate set to $\text{dropout}=0.1$. The optimizer used is AdamW\cite{AdamW}, with a learning rate of $\text{lr}=5e-5$ and a batch size of $\text{bs}=32$.
For the backbone models, we choose Llama2-7B\cite{LLAMA2} and DeepSeek-Coder-7B\cite{DeepSeek-Coder}. \emph{Llama2-7B}\footnote{\url{https://github.com/facebookresearch/llama}} is one of Meta's general pretrained models with fewer parameters, while \emph{DeepSeek-Coder-7B}\footnote{\url{https://github.com/deepseek-ai/deepseek-coder}} is a Chinese and English pretrained model released by DeepSeek AI, pretrained on code and natural language, with a parameter size similar to Llama2-7B.

The experimental results are shown in Table \ref{tab:AgentRE-distillation}, according to which we have the following conclusions. 

1) The models fine-tuned on specific training dataset $D$ perform better than the general models trained on multiple datasets (as shown in Table \ref{tab:AgentRE-main}), such as UIE, USM, etc. It indicates that targeted fine-tuning for specific extraction tasks can achieve better performance compared to multi-task models.

2) The models fine-tuned on the training dataset $D^{\prime}$ containing reasoning rationales perform better than those fine-tuned on $D$, despite the former having significantly less data. It demonstrates that the quality of training data significantly determines the model's performance, and utilizing data derived from the agent's historical reasoning trajectories can better stimulate the reasoning capabilities of smaller models.

3) The experimental results of models trained successively on the two datasets ($D+D^{\prime}$) reveal that, further fine-tuning on the data with reasoning rationales enhances extraction performance for a model already trained on a large amount of simple labelled data.

\section{Conclusion} \label{sec:conclusion}
In this paper, we propose a novel RE framework AgentRE, which effectively leverages various types of information for RE tasks through its retrieval, memory, and extraction modules. 
The experimental results on two representative datasets demonstrates that our AgentRE achieves satisfactory extraction performance in both zero-shot and few-shot unsupervised learning settings, particularly in low-resource scenarios. 
Additionally, ablation and parameter tuning studies confirm the significance of each component of AgentRE for the overall extraction performance. 
Furthermore, AgentRE’s reasoning trajectories can form an effective training dataset containing reasoning rationales, facilitating the transfer of capabilities from larger models to smaller models via distillation learning.
Due to time and cost constraints, our experiments were conducted on only two representative datasets. Future research will include validating the model on more datasets and extending AgentRE to other information extraction tasks.

\begin{acks}
This work was supported by the Chinese NSF Major Research Plan (No.92270121), Youth Fund (No.62102095), Shanghai Science and Technology Innovation Action Plan (No.21511100401). 
\end{acks}

\bibliographystyle{ACM-Reference-Format}
\bibliography{main}

\appendix

\end{document}